\documentclass[11pt,letterpaper]{article}
\usepackage{emnlp2017}
\usepackage{times}
\usepackage{caption}
\usepackage{subcaption}
\usepackage{latexsym}
\usepackage{tikz}
\usepackage{hyperref}
\usepackage{comment}
\usetikzlibrary{arrows}
\usetikzlibrary{positioning,shadows.blur}
\usepackage{pifont}

\usepackage{amsmath,amssymb}
\usepackage{xifthen}
\usepackage{graphicx}
\usepackage{graphics}
\usepackage{float}
\usepackage{algorithm}
\usepackage[noend]{algpseudocode}

\DeclareMathOperator*{\argmax}{\arg\!\max}
\usetikzlibrary{shapes.geometric, arrows, patterns, decorations.pathreplacing}

\newcommand{\them}[0]{\textbf{read:}\ }
\newcommand{\you}[0]{\textbf{write:}\ }

\newcommand{\likelihood}[0]{\textsc{likelihood}\ }
\newcommand{\reinforce}[0]{\textsc{rl}\ }
\newcommand{\rollouts}[0]{\textsc{rollouts}\ }
\newcommand{\rlrollouts}[0]{\textsc{rl+rollouts}\ }

\renewcommand{\vec}[1]{#1}

\emnlpfinalcopy

\def\emnlppaperid{1270}
\usepackage[utf8]{inputenc}

\title{Deal or No Deal? End-to-End Learning for Negotiation Dialogues}

\author{Mike Lewis$^{1}$, Denis Yarats$^{1}$, Yann N. Dauphin$^{1}$, Devi Parikh$^{2,1}$ \and Dhruv Batra$^{2,1}$ \\
  $^{1}${Facebook AI Research} $\qquad$ $^{2}${Georgia Institute of Technology}\\
  {\tt \{mikelewis,denisy,ynd\}@fb.com} $\quad$ {\tt \{parikh,dbatra\}@gatech.edu}}

\date{}

\begin{document}

\maketitle
\begin{abstract}
Much of human dialogue occurs in semi-cooperative settings, where agents with different goals attempt to agree on common decisions.
Negotiations require complex communication and reasoning skills, but success is easy to measure, making this an interesting task for AI.
We gather a large dataset of human-human negotiations on a multi-issue bargaining task, where agents who cannot observe each other's reward functions must reach an agreement (or a deal) via natural 
language dialogue. 
For the first time, we show it is possible to train end-to-end models for negotiation, which must learn both linguistic and reasoning skills with no annotated dialogue states. We also introduce \emph{dialogue rollouts}, in which the model plans ahead by simulating possible complete continuations of the conversation, and find that this technique dramatically improves performance. Our code and dataset are publicly available.\footnote{\url{https://github.com/facebookresearch/end-to-end-negotiator}}


\end{abstract}

\section{Introduction}
\begin{figure*}[ht]
\includegraphics[width=\textwidth]{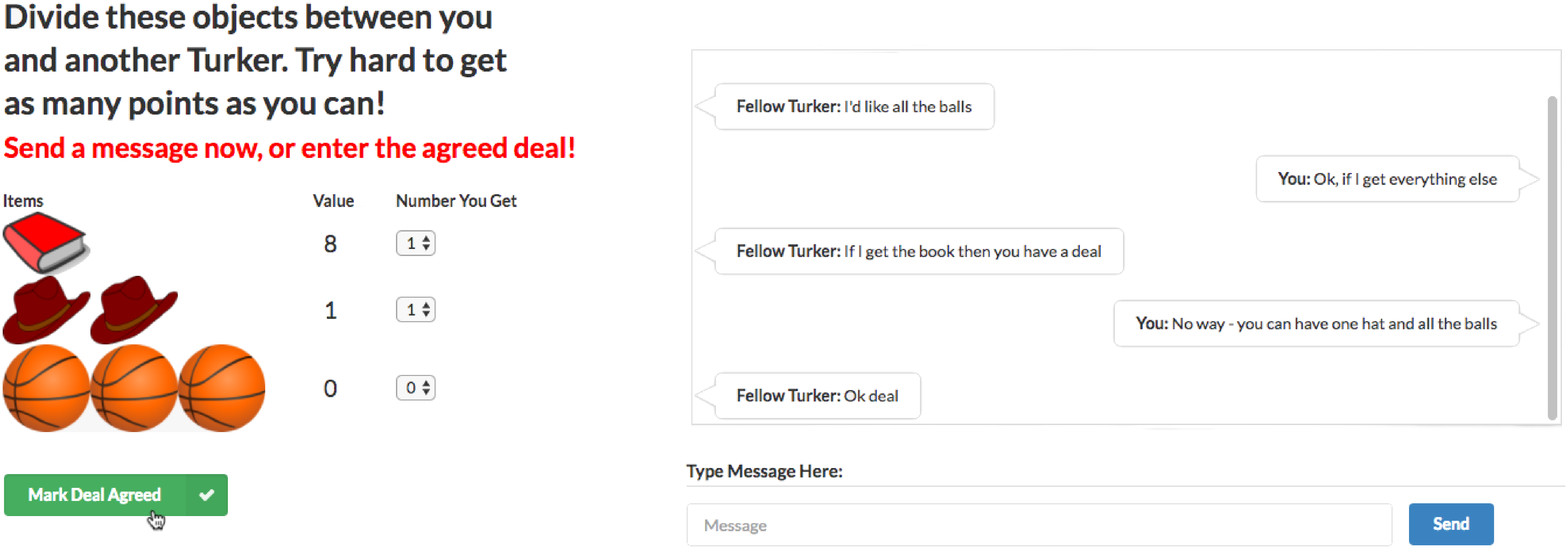}
  \vspace{-4.0em}
\caption{\label{figure:turk}A dialogue in our Mechanical Turk interface, which we used to collect a negotiation dataset.}
\end{figure*}
Intelligent agents often need to cooperate with others who have different goals, and typically use natural language to agree on decisions.
Negotiation is simultaneously a linguistic and a reasoning problem, in which an intent must be formulated and then verbally realised.
Such dialogues contain both cooperative and adversarial elements, and require agents to understand, plan, and generate utterances to achieve their goals \cite{traum:2008,asher:2012}.

We collect the first large dataset of natural language negotiations between two people, and show that end-to-end neural models can be trained to negotiate by maximizing the likelihood of human actions. 
This approach is scalable and domain-independent, but does not model the strategic skills required for negotiating well.
We further show that models can be improved by training and decoding to maximize reward instead of likelihood---by training with self-play reinforcement learning, and using rollouts to estimate the expected reward of utterances during decoding.




To study semi-cooperative dialogue, we gather a dataset of 5808 dialogues between humans on a negotiation task. Users were shown a set of items with a value for each, and asked to agree how to divide the items with another user who has a different, unseen, value function (Figure \ref{figure:turk}). 


We first train recurrent neural networks to imitate human actions.
We find that models trained to maximise the likelihood of human utterances can generate fluent language, but make comparatively poor negotiators, which are overly willing to compromise. We therefore explore two methods for improving the model's strategic reasoning skills---both of which attempt to optimise for the agent's goals, rather than simply imitating humans:

Firstly, instead of training to optimise likelihood, we show that our agents can be considerably improved using \emph{self play}, in which pre-trained models practice negotiating with each other in order to optimise performance.
To avoid the models diverging from human language, we interleave reinforcement learning updates with supervised updates.
For the first time, we show that end-to-end dialogue agents trained using reinforcement learning  outperform their supervised counterparts in negotiations with humans.

Secondly, we introduce a new form of planning for dialogue called \emph{dialogue rollouts}, in which an agent simulates complete dialogues during decoding to estimate the reward of utterances.  We show that decoding to maximise the reward function (rather than likelihood) significantly improves performance against both humans and machines.

Analysing the performance of our agents, we find evidence of sophisticated negotiation strategies.
For example, we find instances of the model feigning interest in a valueless issue, so that it can later `compromise' by conceding it.
Deceit is a complex skill that requires hypothesising the other agent's beliefs, and is learnt relatively late in child development \cite{talwar:2002}. Our agents have \emph{learnt to deceive} without any explicit human design, simply by trying to achieve their goals.



The rest of the paper proceeds as follows: \S\ref{section:data} describes the collection of a large dataset of human-human negotiation dialogues.
\S\ref{section:likelihood} describes a baseline supervised model, which we then show can be improved by goal-based training (\S\ref{section:reinforcement_learning}) and decoding (\S\ref{section:decoding}).  \S\ref{section:experiments} measures the performance of our models and humans on this task, and \S\ref{section:analysis} gives a detailed analysis and suggests future directions.





\section{Data Collection}
\begin{figure*}[ht!]
\newcommand{\turnA}{I want the books and the hats, you get the ball}
\newcommand{\turnB}{Give me a book too and we have a deal}
\newcommand{\turnC}{Ok, deal}
\newcommand{\turnD}{\textless choose\textgreater}
\newcommand{\choiceA}{2x\textit{book} 2x\textit{hat}}
\newcommand{\choiceB}{1x\textit{book} 1x\textit{ball}}

\newcommand{\itemvalue}[3]{#2x\textit{#1} & \textit{value}=#3}
\newcommand{\bookA}{\itemvalue{book}{3}{1}}
\newcommand{\hatA}{\itemvalue{hat}{2}{3}}
\newcommand{\ballA}{\itemvalue{ball}{1}{1}}

\newcommand{\bookB}{\itemvalue{book}{3}{2}}
\newcommand{\hatB}{\itemvalue{hat}{2}{1}}
\newcommand{\ballB}{\itemvalue{ball}{1}{2}}

\newcommand{\goals}[3]{
\setlength{\tabcolsep}{0.2em} 
\begin{tabular}{ll}
     #1\\#2\\#3
\end{tabular}
}

\tikzstyle{arrow} = [thick,->]
\tikzset{>=latex}

\begin{tikzpicture} 
  \tikzstyle{box} = [rectangle, rounded corners, font=\sffamily\footnotesize, text width=2.45cm, draw=black]

      \node [box, draw=black, thick, fill = white, minimum width = 6.7cm, minimum height = 8cm]  at (current page.north west) (raw)
    {};
    
    \node[below right] at (raw.north west) (rawlabel) {\underline{\bfseries Crowd Sourced Dialogue
}};

    \node [box, fill=blue!30, text width      = 2.45cm, below=0.0cm of rawlabel.south west, anchor=north west, xshift=0.2cm]
    (User1Goals)
    {\underline{\bfseries Agent 1 Input}\\
      \goals{\bookA}{\hatA}{\ballA}   
     };
     
    \node [box, right = 0.8cm of User1Goals, fill = blue!30, text width      = 2.45cm]
    (User2Goals)
    {\underline{\bfseries Agent 2 Input}\\
      \goals{\bookB}{\hatB}{\ballB}   
    };
      
    \node [box, below = 0.9cm of User1Goals.south west, anchor=north west, fill = yellow, text width      = 6cm]
    (Dialog)
    {\underline{\bfseries Dialogue}\\
      \textbf{Agent 1:} \turnA\\
      \textbf{Agent 2:} \turnB\\
      \textbf{Agent 1:} \turnC\\
      \textbf{Agent 2:} \turnD\\

    }; 
    
    \node [box, below = 0.8cm of Dialog.south west, anchor=north west, fill = red!30, text width=2.3cm]
    (User1Choice)
    {\underline{\bfseries Agent 1 Output}\\
      \choiceA\\      
    };

    \node [box, right = 1.2cm of User1Choice, fill = red!30, text width=2.3cm]
    (User2Choice)
    {\underline{\bfseries Agent 2 Output}\\
      \choiceB\\      
    };

    \node [box, right = 1cm of raw.north east, ,anchor=north west, draw=black, thick, fill = white, minimum width = 8cm, minimum height = 3.75cm] (perspective1)
    {};

    \draw [arrow] (User1Goals) -- (Dialog);
    \draw [arrow] (User2Goals) -- (Dialog);
    \draw [arrow] (Dialog) -- (User1Choice);
    \draw [arrow] (Dialog) -- (User2Choice);

    \node[below right] at (perspective1.north west) (perspective1label) {\underline{\bfseries Perspective: Agent 1}};

    \node [box, below = 0.5cm of perspective1, draw=black, thick, fill = white, minimum width = 8cm, minimum height = 3.75cm] (perspective2)
    {};
    
    \node[below right] at (perspective2.north west) (perspective2label) {\underline{\bfseries Perspective: Agent 2}};

    \node [box, fill = blue!30, text width      = 2.45cm, below=0.0cm of perspective1label.south west, anchor=north west, xshift=0.2cm]
    (perspective1_goals)
    {\underline{\bfseries Input}\\
      \goals{\bookA}{\hatA}{\ballA}   
     };
     
         \node [box, fill = red!30, text width      = 2.45cm, below = 0.2cm of perspective1_goals]
    (perspective1_choice)
    {\underline{\bfseries Output}\\
      \choiceA\\      
     };

    \node [box, right = 1cm of perspective1_goals.north east, anchor=north west, fill = yellow, text width      = 3.5cm]
    (perspective1_dialog)
    {\underline{\bfseries Dialogue}\\
      \textbf{write:} \turnA\ \textbf{read:} \turnB\ 
      \textbf{write:} \turnC\ \textbf{read:} \turnD
    }; 
    
        \node [box, fill = blue!30, text width      = 2.45cm, below=0.0cm of perspective2label.south west, anchor=north west, xshift=0.2cm]
    (perspective2_goals)
    {\underline{\bfseries Input}\\
      \goals{\bookB}{\hatB}{\ballB}   
     };
    
        \node [box, right = 1cm of perspective2_goals.north east, anchor=north west, fill = yellow, text width      = 3.5cm]
    (perspective2_dialog)
    {\underline{\bfseries Dialogue}\\
      \textbf{read:} \turnA\ 
      \textbf{write:} \turnB\ 
      \textbf{read:} \turnC\ 
      \textbf{write:} \turnD
    }; 
    
             \node [box, fill = red!30, text width      = 2.45cm, below = 0.2cm of perspective2_goals]
    (perspective2_choice)
    {\underline{\bfseries Output}\\
      \choiceB\\      
     };
    
    \draw [arrow] (perspective1_goals) -- (perspective1_dialog);
    \draw [arrow] (perspective2_goals) -- (perspective2_dialog);

    \draw [arrow] (perspective1_dialog) -- (perspective1_choice.east);
    \draw [arrow] (perspective2_dialog) -- (perspective2_choice.east);
\end{tikzpicture}
\caption{\label{figure:representation} Converting a crowd-sourced dialogue (left) into two training examples (right), from the perspective of each user. The perspectives differ on their input goals, output choice, and in special tokens marking whether a statement was read or written. We train conditional language models to predict the dialogue given the input, and additional models to predict the output given the dialogue. }
\end{figure*}
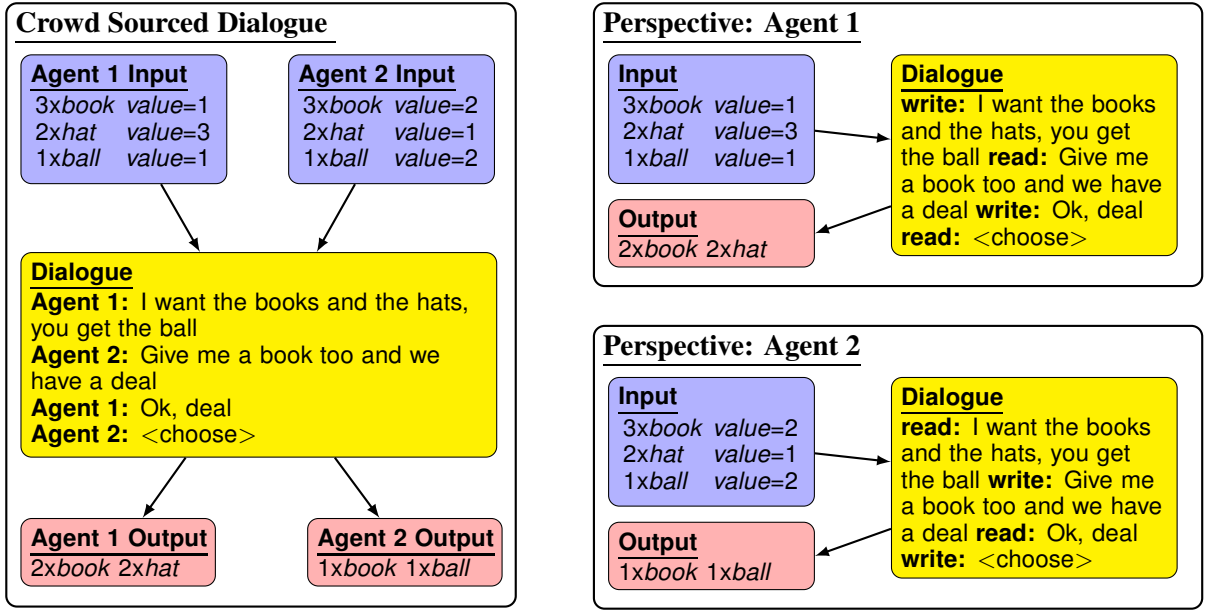

\label{section:data}
\subsection{Overview}
To enable end-to-end training of negotiation agents, we first 
develop a novel negotiation task and 
curate a dataset of human-human dialogues for this task. This task and dataset follow our proposed general framework for studying semi-cooperative dialogue.
Initially, each agent is shown an input specifying a space of possible actions and a reward function which will score the outcome of the negotiation.
Agents then sequentially take turns of either sending natural language messages, or selecting that a final decision has been reached.
When one agent selects that an agreement has been made, both agents independently output what they think the agreed decision was.
If conflicting decisions are made, both agents are given zero reward.





\subsection{Task}
Our task is an instance of \emph{multi issue bargaining} \cite{fershtman:1990}, and is based on \newcite{devault:2015}. 
Two agents are both shown the same collection of items, and instructed to divide them so that each item assigned to one agent.

Each agent is given a different randomly generated value function, which gives a non-negative value for each item. 
The value functions are constrained so that: (1) the total value for a user of all items is 10; (2) each item has non-zero value to at least one user; and (3) some items have non-zero value to both users. These constraints enforce that it is not possible for both agents to receive a maximum score, and that no item is worthless to both agents, so the negotiation will be competitive.
After 10 turns, we allow agents the option to complete the negotiation with no agreement, which is worth 0 points to both users.
We use 3 item types (\emph{books}, \emph{hats}, \emph{balls}), and between 5 and 7 total items in the pool.
Figure \ref{figure:turk} shows our interface.

\subsection{Data Collection}
We collected a set of human-human dialogues using Amazon Mechanical Turk. 
Workers were paid \$0.15 per dialogue, with a \$0.05 bonus for maximal scores.
We only used workers based in the United States with a 95\% approval rating and at least 5000 previous HITs.
Our data collection interface was adapted from that of \newcite{das:2016}.

We collected a total of 5808 dialogues, based on 2236 unique scenarios (where a scenario is the available items and values for the two users). We held out a test set of 252 scenarios (526 dialogues). 
Holding out test scenarios means that models must generalise to new situations.

\section{Likelihood Model}
\def\layersep{1.0cm}
\newcommand{\counts}[2]{\textbf{#1} \textit{count}=#2}
\newcommand{\values}[2]{\textbf{#1} \textit{value}=#2}

\def\goals{}

    \tikzstyle{every pin edge}=[<-,shorten <=1pt]
    \tikzstyle{neuron}=[circle,fill=black!25,minimum size=17pt,inner sep=0pt]
    \tikzstyle{input neuron}=[neuron,draw=black!50, fill=white]
    \tikzstyle{output neuron}=[neuron, fill=red!50]
    \tikzstyle{hidden neuron}=[neuron, fill=blue!50]
    \tikzstyle{annot} = [text width=4em, text centered]
  \tikzstyle{box} = [rectangle, rounded corners, font=\sffamily\footnotesize, text width=2.3cm, draw=black]

\newcommand{\network}[0]{
\scalebox{0.7}{
\begin{tikzpicture}[shorten >=1pt,->,draw=black!50, node distance=\layersep]
\tikzset{>=latex}

      \node [box, draw=black, thick, fill = white
      ] at (2,0.2) (enc) {Input Encoder};

      \node [box, draw=black, thick, fill = white
      , right = 1cm of enc]  (dec) {Output Decoder};

    \foreach[count=\y,evaluate=\y as \prevx using int(\y-2)] \name in \words {
        \node[input neuron] (H-\y) at (\prevx, -2) {};
        \node[anchor=base] (O-\y) at (\prevx, -3) {\name};
    }

    \foreach[count=\y,evaluate=\y as \prevx using int(\y-1)] \w in \written {
        \ifthenelse{\y>1}{\path (O-\prevx) edge (H-\y);}{}
        \ifthenelse{\y>1}{\path (H-\prevx) edge (H-\y);}{}
        \ifthenelse{\y>0 \AND \w=1}{\path (H-\y) edge (O-\y);}{}

        \path (enc) edge (H-\y);
        \path (H-\y) edge (dec);
    }

\end{tikzpicture}
}
}

\begin{figure*}[t!]
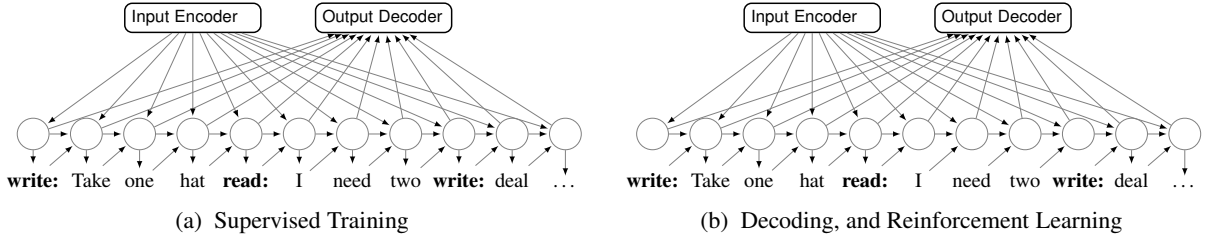

    \def\words{\you,\vphantom{.}\space\space Take, one, hat, \them, I, need, two, \you, deal, \dots}

    \centering
    \begin{subfigure}[t]{0.5\textwidth}
        \def\written{1, 1, 1, 1, 1, 1, 1, 1, 1, 1, 1}
        \centering
        \network
        \caption{\label{figure:model:supervised} Supervised Training}
    \end{subfigure}%
    ~ 
    \begin{subfigure}[t]{0.5\textwidth}
        \def\written{0, 1, 1, 1, 1, 0, 0, 0, 0, 1, 1}
        \centering
        \network
        \caption{\label{figure:model:reinforcement} Decoding, and Reinforcement Learning}
    \end{subfigure}
    \caption{\label{figure:model} Our model: tokens are predicted conditioned on previous words and the input, then the output is predicted using attention over the complete dialogue. In supervised training (\ref{figure:model:supervised}), we train the model to predict the tokens of \emph{both} agents. During decoding and reinforcement learning (\ref{figure:model:reinforcement}) some tokens are sampled from the model, but some are generated by the other agent and are only encoded by the model. }
\end{figure*}

\label{section:likelihood}
We propose a simple but effective baseline model for the conversational agent, in which a sequence-to-sequence model is trained to produce the complete dialogue, conditioned on an agent's input. 


\subsection{Data Representation}
\label{section:data_representation}
Each dialogue is converted into two training examples, showing the complete conversation from the perspective of each agent.
The examples differ on their input goals, output choice, and whether utterances were read or written.

Training examples contain an input goal $\vec{g}$, specifying the available items and their values, a dialogue $\vec{x}$, and an output decision $\vec{o}$ specifying which items each agent will receive. 
Specifically, we represent $\vec{g}$ as a list of six integers corresponding to the count and value of each of the three item types.
Dialogue $\vec{x}$ is a list of tokens $x_{0..T}$ containing the turns of each agent interleaved with symbols marking whether a turn was written by the agent or their partner, terminating in a special token indicating one agent has marked that an agreement has been made.
Output $\vec{o}$ is six integers describing how many of each of the three item types are assigned to each agent.
See Figure \ref{figure:representation}.

\subsection{Supervised Learning}
\label{section:supervised}

We train a sequence-to-sequence network to generate an agent's perspective of the dialogue conditioned on the agent's input goals (Figure \ref{figure:model:supervised}).

The model uses 4 recurrent neural networks, implemented as GRUs \cite{cho:2014}: $\text{GRU}_w$, $\text{GRU}_g$,  $\text{GRU}_{\overrightarrow{o}}$, and $\text{GRU}_{\overleftarrow{o}}$.

The agent's input goals $\vec{g}$ are encoded using $\text{GRU}_g$. We refer to the final hidden state as $h^g$. The model then predicts each token $x_t$ from left to right, conditioned on the previous tokens and $h^g$. At each time step $t$, $\text{GRU}_w$ takes as input the previous hidden state $h_{t-1}$, previous token $x_{t-1}$ (embedded with a matrix $E$), and input encoding $h^g$. Conditioning on the input at each time step helps the model learn dependencies between language and goals.
\begin{equation}
h_t = \text{GRU}_w(h_{t-1}, [Ex_{t-1}, h^g])
\end{equation}
The token at each time step is predicted with a softmax, which uses weight tying with the embedding matrix $E$ \cite{mao:2015}:
\begin{equation}
p_\theta (x_t | x_{0.. t-1}, \vec{g}) \propto \exp(E^T h_t)
\end{equation}
Note that the model predicts both agent's words, enabling its use as a forward model in Section  \ref{section:decoding}.

At the end of the dialogue, the agent outputs a set of tokens $\vec{o}$ representing the decision. We generate each output conditionally independently, using a separate classifier for each. The classifiers share bidirectional $\text{GRU}_o$ and attention mechanism \cite{bahdanau:2014} over the dialogue, and additionally conditions on the input goals.
\begin{align}
h^{\overrightarrow{o}}_t &= \text{GRU}_{\overrightarrow{o}}(h^{\overrightarrow{o}}_{t-1}, [Ex_{t}, h_t])\\
h^{\overleftarrow{o}}_t &= \text{GRU}_{\overleftarrow{o}}(h^{\overleftarrow{o}}_{t+1}, [Ex_{t}, h_t])\\
h^o_t&=[h^{\overleftarrow{o}}_t, h^{\overrightarrow{o}}_t]\\
h^a_t&=W[\tanh(W'h^o_t)]\\
\alpha_t &= \frac {\exp(w \cdot h^a_t)}{\sum_{t'} \exp(w\cdot h^a_{t'})}\\
h^s &= \tanh(W^s[h^g, \sum_t \alpha_t h_t])
\end{align}
 
The output tokens are predicted using softmax:
\begin{equation}
p_\theta(o_i | x_{0..t}, \vec{g}) \propto \exp(W^{o_i} h^s)
\end{equation}

The model is trained to minimize the negative log likelihood of the token sequence $x_{0..T}$ conditioned on the input goals $\vec{g}$, and of the outputs $\vec{o}$ conditioned on $\vec{x}$ and $\vec{g}$. The two terms are weighted with a hyperparameter $\alpha$.

\begin{align}
\label{eq:supervised}
L(\theta) =& -\underbrace{\sum_{\vec{x},\vec{g}} \sum_{t} \log p_\theta(x_t | x_{0.. t-1}, \vec{g})}_{\text{Token prediction loss}}\nonumber\\
          & -\alpha \underbrace{\sum_{\vec{x},\vec{g},\vec{o}} \sum_{j} \log p_\theta(o_j | x_{0.. T}, \vec{g})}_{\text{Output choice prediction loss}}
\end{align}


Unlike the Neural Conversational Model \cite{vinyals:2015}, our approach shares all parameters for reading and generating tokens.

\subsection{Decoding}
\label{section:likelihood_decoding}
During decoding, the model must generate an output token $x_t$ conditioned on dialogue history $x_{0..t-1}$ and input goals $\vec{g}$, by sampling from $p_\theta$:

\begin{equation}
 x_t \sim p_\theta(x_t | x_{0..t-1}, \vec{g})
\end{equation}

If the model generates a special \emph{end-of-turn} token, it then encodes a series of tokens output by the other agent, until its next turn (Figure \ref{figure:model:reinforcement}).

The dialogue ends when either agent outputs a special \emph{end-of-dialogue} token. The model then outputs a set of choices $\vec{o}$. We choose each item independently, but enforce consistency by checking the solution is in a feasible set $O$:
\begin{equation}
\label{equation:output}
\vec{o}^{*} = \argmax_{\vec{o}\in O} \prod_{i} p_\theta(o_i | x_{0.. T}, \vec{g})
\end{equation}
In our task, a solution is feasible if each item is assigned to exactly one agent. The space of solutions is small enough to be tractably enumerated.


\section{Goal-based Training}
\label{section:reinforcement_learning}
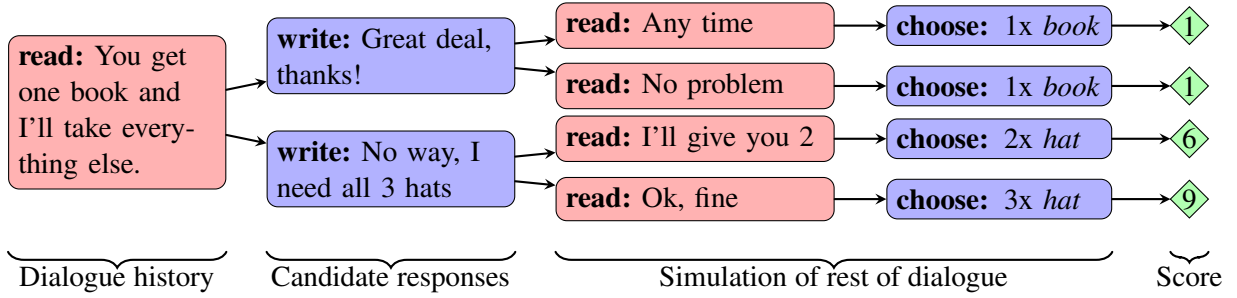
\begin{figure*}[ht]


\begin{tikzpicture}[scale=.8,cap=round]
\tikzstyle{them} = [rectangle, rounded corners, minimum width=2cm, text width=2.6cm, minimum height=1cm, draw=black, fill=red!30]
\tikzstyle{candidate} = [rectangle, rounded corners, minimum width=2cm, text width=3cm, minimum height=0.3cm, draw=black, fill=blue!30]
\tikzstyle{rollout} = [rectangle, rounded corners, minimum width=3.4cm, minimum height=0.3cm, text width=3.4cm, draw=black, fill=red!30]
\tikzstyle{score} = [diamond, minimum width=0.5cm, minimum height=0.5cm, draw=black, fill=green!30, inner sep=0,outer sep=0]

\tikzstyle{arrow} = [thick,->,>=stealth]

\node (start) [them] {\them You get one book and I'll take everything else.};
\node (candidate2) [candidate, right of=start, xshift=2.6cm, yshift=0.75cm] {
\you Great deal, thanks!};
\node (candidate1) [candidate, right of=start, xshift=2.6cm, yshift=-0.75cm] {\you No way, I need all 3 hats};

\node (rollout11) [rollout, right of=candidate1, xshift=3cm, yshift=-0.4cm] {\them Ok, fine};
\node (rollout12) [rollout, right of=candidate1, xshift=3cm, yshift=0.4cm] {\them I'll give you 2};

\node (rollout21) [rollout, right of=candidate2, xshift=3cm, yshift=-0.4cm] {\them No problem};
\node (rollout22) [rollout, right of=candidate2, xshift=3cm, yshift=0.4cm] {\them Any time};

\node (choose11) [candidate, right of=rollout11, xshift=3cm, text width=2.7cm] {\textbf{choose:} 3x \textit{hat}};
\node (choose12) [candidate, right of=rollout12, xshift=3cm, text width=2.7cm] {\textbf{choose:} 2x \textit{hat}};

\node (choose21) [candidate, right of=rollout21, xshift=3cm, text width=2.7cm] {\textbf{choose:} 1x \textit{book}};
\node (choose22) [candidate, right of=rollout22, xshift=3cm, text width=2.7cm] {\textbf{choose:} 1x \textit{book}};

\node (score11) [score, right of=choose11, xshift=1.5cm] {9};
\node (score12) [score, right of=choose12, xshift=1.5cm] {6};

\node (score21) [score, right of=choose21, xshift=1.5cm] {1};
\node (score22) [score, right of=choose22, xshift=1.5cm] {1};

\draw [arrow] (start) -- (candidate1);
\draw [arrow] (start) -- (candidate2);

\foreach \i in {1,2}
{
\draw [arrow] (candidate1) -- (rollout1\i);
\draw [arrow] (candidate2) -- (rollout2\i);

\foreach \j in {1,2}
{
\draw [arrow] (rollout\i\j) -- (choose\i\j);
\draw [arrow] (choose\i\j) -- (score\i\j);
}
}

\draw [
    thick,
    decoration={
        brace,
        mirror,
        raise=1.85cm,amplitude=6pt
    },
    decorate
] (start.west) -- (start.east) node [pos=0.5,anchor=north,yshift=-1.9cm] {Dialogue history}; 

\draw [
    thick,
    decoration={
        brace,
        mirror,
        raise=1.10cm,amplitude=6pt
    },
    decorate
] (candidate1.west) -- (candidate1.east) node [pos=0.5,anchor=north,yshift=-1.15cm] {Candidate responses}; 

\draw [
    thick,
    decoration={
        brace,
        mirror,
        raise=1.5cm,amplitude=6pt
    },
    decorate
] (rollout12.west) -- (choose12.east) node [pos=0.5,anchor=north,yshift=-1.55cm] {Simulation of rest of dialogue}; 

\draw [
    thick,
    decoration={
        brace,
        mirror,
        raise=1.5cm,amplitude=4pt
    },
    decorate
] (score12.west) -- (score12.east) node [pos=0.5,anchor=north,yshift=-1.55cm] {Score};

\end{tikzpicture}
\caption{
\label{fig:rollouts} 
Decoding through rollouts: The model first generates a small set of candidate responses. For each candidate it simulates the future conversation by sampling, and estimates the expected future reward by averaging the scores. The system outputs the candidate with the highest expected reward.
}\vspace{-2.5mm}
\end{figure*}

Supervised learning aims to imitate the actions of human users, but does not explicitly attempt to maximise an agent's goals.
Instead, we explore pre-training with supervised learning, and then fine-tuning against the evaluation metric using reinforcement learning. Similar two-stage learning strategies have been used previously (e.g. \newcite{li:2016, das:2017}).


During reinforcement learning, an agent $A$ attempts to improve its parameters from conversations with another agent $B$. While the other agent $B$ could be a human, in our experiments we used our fixed supervised model that was trained to imitate humans. The second model is fixed as we found that updating the parameters of both agents led to divergence from human language. In effect, agent $A$ learns to improve by simulating conversations with the help of a surrogate forward model.

Agent $A$ reads its goals $\vec{g}$ and then generates tokens $x_{0.. n}$ by sampling from $p_\theta$. When $x$ generates an end-of-turn marker, it then reads in tokens $x_{n+1.. m}$ generated by agent $B$. These turns alternate until one agent emits a token ending the dialogue. Both agents then output a decision $\vec{o}$ and collect a reward from the environment (which will be 0 if they output different decisions). We denote the subset of tokens generated by $A$ as $X^A$ (e.g. tokens with incoming arrows in Figure \ref{figure:model:reinforcement}).


After a complete dialogue has been generated, we update agent $A$'s parameters based on the outcome of the negotiation.
Let $r^A$ be the score agent $A$ achieved in the completed dialogue, $T$ be the length of the dialogue, $\gamma$ be a discount factor that rewards actions at the end of the dialogue more strongly, and $\mu$ be a running average of completed dialogue rewards so far\footnote{As all rewards are non-negative, we instead re-scale them by subtracting the mean reward found during self play. Shifting in this way can reduce the variance of our estimator.}. We define the future reward $R$ for an action $x_t\in X^A$ as follows:
\begin{equation}
R(x_t) = \sum_{x_t\in X^{A}}\gamma^{T-t}(r^A(\vec{o}) - \mu)
\end{equation}

We then optimise the expected reward of each action $x_t\in X^A$:

\begin{equation}
L^{RL}_\theta = \mathbb{E}_{x_{t}\sim p_\theta(x_{t}|x_{0..t-1},\vec{g})}[R(x_t)]
\end{equation}

The gradient of $L^{RL}_\theta$ is calculated as in \textsc{REINFORCE} \cite{williams:1992}:

\begin{equation}
\nabla_\theta L^{RL}_\theta = \!\!\!\!\!\sum_{x_t\in X^{A}}\!\!\!\mathbb{E}_{x_t}[R(x_t)\nabla_\theta \log(p_\theta(x_{t}|x_{0..t-1},\vec{g}))]
\end{equation}

\section{Goal-based Decoding}
\setlength{\textfloatsep}{10pt}
\begin{algorithm}[t]

\begin{algorithmic}[1]
\algdef{SE}[DOWHILE]{Do}{doWhile}{\algorithmicdo}[1]{\algorithmicwhile\ #1}%
\algrenewcommand\algorithmicindent{1.0em}%
\makeatletter
\algnewcommand{\LineComment}[1]{\Statex \hskip\ALG@thistlm \(\triangleright\) #1}
\makeatother

\Procedure{Rollout}{$x_{0.. i}, \vec{g}$} 
    \State $\vec{u}^* \gets \varnothing$
    \For{$c\in\{1..C\}$} \Comment{$C$ candidate moves}
      \State{$j\gets i$}
      \Do \Comment{Rollout to end of turn}
        \State $j\gets j+1$
        \State $x_{j} \sim p_\theta(x_{j}| x_{0..j-1}, \vec{g}) $
      \doWhile{$x_k\notin \{\text{\emph{read:}, \emph{choose:}}\}$}
      \State $u \gets x_{i+1}..x_{j}$ \Comment{$u$ is candidate move}
      \For{$s \in \{1..S\}$}
        \Comment{$S$ samples per move}
        \State $k\gets j$ \Comment{Start rollout from end of $u$}
        \While{$x_k\not=\text{\emph{choose:}}$}
            \LineComment{Rollout to end of dialogue}
            \State $k\gets k+1$
            \State $x_{k} \sim p_\theta(x_{k}| x_{0..k-1}, \vec{g}) $
        \EndWhile
        \LineComment{Calculate rollout output and reward}
        \State $\vec{o}\gets \argmax_{\vec{o'}\in O} p(\vec{o'} | x_{0..k}, \vec{g})$
        \State $R(u) \gets R(u) + r(\vec{o})p(\vec{o'} | x_{0..k}, \vec{g})$
      \EndFor
      \If{$R(u) > R({\vec{u}^*})$}
        \State $\vec{u}^* \gets u$
      \EndIf
    \EndFor
      \State \textbf{return} $\vec{u}^*$ \Comment{Return best move}
\EndProcedure
\end{algorithmic}
\caption{Dialogue Rollouts algorithm.\label{algorithm:rollouts}}
\

\end{algorithm}

\label{section:decoding}
Likelihood-based decoding (\S \ref{section:likelihood_decoding}) may not be optimal. For instance, an agent may be choosing between accepting an offer, or making a counter offer.
The former will often have a higher likelihood under our model, as there are fewer ways to agree than to make another offer, but the latter may lead to a better outcome.
Goal-based decoding also allows more complex dialogue strategies. For example, a deceptive utterance is likely to have a low model score (as users were generally honest in the supervised data), but may achieve high reward.

We instead explore decoding by maximising expected reward.
We achieve this by using $p_\theta$ as a forward model for the complete dialogue, and then deterministically computing the reward. Rewards for an utterance are averaged over samples to calculate expected future reward (Figure \ref{fig:rollouts}).

We use a two stage process: First, we generate $c$ candidate utterances $U=u_{0..c}$, representing possible complete turns that the agent could make, which are generated by sampling from $p_\theta$ until the \emph{end-of-turn} token is reached.  Let $x_{0..n-1}$ be current dialogue history. We then calculate the expected reward $R(u)$ of candidate utterance $u=x_{n,n+k}$ by repeatedly sampling $x_{n+k+1,T}$ from $p_\theta$, then choosing the best output $\vec{o}$ using Equation \ref{equation:output}, and finally deterministically computing the reward $r(\vec{o})$. The reward is scaled by the probability of the output given the dialogue, because if the agents select different outputs then they both receive 0 reward.
\begin{align}
R(x_{n.. n+k})=\mathbb{E}_{x_{(n+k+1.. T;\vec{o})}\sim p_\theta}[r(\vec{o})p_\theta(\vec{o}|x_{0..T})]
\end{align}

We then return the utterance maximizing $R$.
\begin{equation}
u^{*}=\argmax_{u\in U}R(u)
\end{equation}
We use 5 rollouts for each of 10 candidate turns.




\section{Experiments}
\label{section:experiments}
\begin{table*}
\centering
\scalebox{0.85}{\begin{tabular}{c|cccc|cccc|}
\cline{2-9} & \multicolumn{4}{c|}{vs. \likelihood} & \multicolumn{4}{c|}{vs. Human} \\ 
\hline
\multicolumn{1}{|c|}{Model} & \begin{tabular}[c]{@{}c@{}}Score \\ (all)\end{tabular} & \begin{tabular}[c]{@{}c@{}}Score \\ (agreed)\end{tabular}& \begin{tabular}[c]{@{}c@{}}\% \\ Agreed\end{tabular} & \begin{tabular}[c]{@{}c@{}}\% Pareto \\ Optimal\end{tabular} & \begin{tabular}[c]{@{}c@{}}Score \\ (all)\end{tabular} & \begin{tabular}[c]{@{}c@{}}Score \\ (agreed)\end{tabular} & \begin{tabular}[c]{@{}c@{}}\% \\ Agreed\end{tabular}& \begin{tabular}[c]{@{}c@{}}\% Pareto\\ Optimal\end{tabular} \\
\hline
\multicolumn{1}{|c|}{\likelihood} & \multicolumn{1}{c}{5.4 vs. 5.5} & \multicolumn{1}{c}{6.2 vs. 6.2} & 
\multicolumn{1}{c}{87.9}& \multicolumn{1}{c|}{49.6} & \multicolumn{1}{c}{4.7 vs. 5.8} & 
\multicolumn{1}{c}{6.2 vs. 7.6} & 
\multicolumn{1}{c}{\textbf{76.5}} &
\multicolumn{1}{c|}{66.2} \\

\multicolumn{1}{|c|}{\reinforce} & \multicolumn{1}{c}{7.1 vs. 4.2} & \multicolumn{1}{c}{7.9 vs. 4.7} & \multicolumn{1}{c}{89.9}& \multicolumn{1}{c|}{58.6} & \multicolumn{1}{c}{4.3 vs. 5.0} & \multicolumn{1}{c}{6.4 vs. 7.5} & \multicolumn{1}{c}{67.3} &\multicolumn{1}{c|}{69.1} \\

\multicolumn{1}{|c|}{\rollouts} & \multicolumn{1}{c}{7.3 vs. 5.1} & \multicolumn{1}{c}{7.9 vs. 5.5} & \multicolumn{1}{c}{92.9}& \multicolumn{1}{c|}{63.7} & \multicolumn{1}{c}{\textbf{5.2 vs. 5.4}} & \multicolumn{1}{c}{7.1 vs. 7.4} & \multicolumn{1}{c}{72.1} &\multicolumn{1}{c|}{78.3} \\

\multicolumn{1}{|c|}{\rlrollouts}  & \multicolumn{1}{l}{\textbf{8.3 vs. 4.2}} & \multicolumn{1}{c}{\textbf{8.8 vs. 4.5}} & \multicolumn{1}{c}{\textbf{94.4}} & \multicolumn{1}{c|}{\textbf{74.8}} & \multicolumn{1}{c}{4.6 vs. 4.2} & \multicolumn{1}{c}{\textbf{8.0 vs. 7.1}} &\multicolumn{1}{c}{57.2} & \multicolumn{1}{c|}{\textbf{82.4}} \\
\hline
\end{tabular}}

\caption{\label{table:endtoend} End task evaluation on heldout scenarios, against the \likelihood model and humans from Mechanical Turk. The maximum score is 10. \emph{Score (all)} gives 0 points when agents failed to agree.}\vspace{-3.5mm}
\end{table*}

\begin{table}
\centering
\label{my-label}
\scalebox{0.80}{\begin{tabular}{|c|c|} 
\hline
Metric & Dataset \\ 
\hline
Number of Dialogues & 5808  \\
Average Turns per Dialogue  &  6.6 \\
Average Words per Turn      &  7.6 \\
\% Agreed           & 80.1 \\
Average Score (/10) &  6.0 \\
\% Pareto Optimal   & 76.9 \\
\hline
\end{tabular}}
\caption{\label{tab:data} Statistics on our dataset of crowd-sourced dialogues between humans.}
\end{table}

\subsection{Training Details}
We implement our models using PyTorch.
All hyper-parameters were chosen on a development dataset. 
The input tokens are embedded into a 64-dimensional space, while the dialogue tokens are embedded with 256-dimensional embeddings (with no pre-training). 
The input $GRU_g$ has a hidden layer of size 64 and the dialogue $GRU_w$ is of size 128.
The output $\text{GRU}_{\overrightarrow{o}}$ and $\text{GRU}_{\overleftarrow{o}}$ both have a hidden state of size 256, the size of $h^s$ is 256 as well.
During supervised training, we optimise using stochastic gradient descent with a minibatch size of 16, an initial learning rate of 1.0, Nesterov momentum  with $\mu$=0.1 \cite{nesterov:1983}, and clipping gradients whose $L^2$ norm exceeds 0.5.
We train the model for 30 epochs and pick the snapshot of the model with the best validation perplexity. We then annealed the learning rate by a factor of 5 each epoch.
We weight the terms in the loss function (Equation \ref{eq:supervised}) using $\alpha$=0.5. We do not train against output decisions where humans selected different agreements.
Tokens occurring fewer than 20 times are replaced with an `unknown' token. 

During reinforcement learning, we use a learning rate of 0.1, clip gradients above 1.0, and use a discount factor of $\gamma$=0.95. After every 4 reinforcement learning updates, we make a supervised update with mini-batch size 16 and learning rate 0.5, and we clip gradients at 1.0. We used 4086 simulated conversations.

When sampling words from $p_\theta$, we reduce the variance by doubling the values of logits (i.e. using temperature of 0.5).

\subsection{Comparison Systems}
We compare the performance of the following: \likelihood uses supervised training and decoding  (\S\ref{section:likelihood}), \reinforce is fine-tuned with goal-based self-play (\S\ref{section:reinforcement_learning}), \rollouts uses supervised training combined with goal-based decoding using rollouts (\S\ref{section:decoding}), and \rlrollouts uses rollouts with a base model trained with reinforcement learning.

\subsection{Intrinsic Evaluation}
For development, we use measured the perplexity of user generated utterances, conditioned on the input and previous dialogue.

\begin{table}
\centering
\scalebox{0.80}{\begin{tabular}{|c|c|c|c|} 
\hline
Model & Valid PPL & Test PPL & Test Avg. Rank \\
\hline
\likelihood & 5.62 & 5.47 & 521.8 \\
\reinforce &  6.03 & 5.86   &  517.6  \\
\rollouts  & - & -   & 844.1 \\
\rlrollouts & - & - & 859.8 \\
\hline
\end{tabular}}
\caption{\label{table:intrinsic} Intrinsic evaluation showing the average perplexity of tokens and rank of complete turns (out of 2083 unique human messages from the test set). Lower is more human-like for both. 
}
\end{table}


Results are shown in Table \ref{table:intrinsic}, and show that the simple \likelihood model produces the most human-like responses, and the alternative training and decoding strategies cause a divergence from human language. 
Note however, that this divergence may not necessarily correspond to lower quality language---it may also indicate different strategic decisions about what to say. Results in \S\ref{section:endtoend} show all models could converse with humans.

\subsection{End-to-End Evaluation}
\label{section:endtoend}
We measure end-to-end performance in dialogues both with the likelihood-based agent and with humans on Mechanical Turk, on held out scenarios. 

Humans were told that they were interacting with other humans, as they had been during the collection of our dataset (and few appeared to realize they were in conversation with machines).

We measure the following statistics:\\
\textbf{Score:} The average score for each agent (which could be a human or model), out of 10.
\\ \textbf{Agreement:} The percentage of dialogues where both agents agreed on the same decision.\\ \textbf{Pareto Optimality:} The percentage of Pareto optimal solutions for agreed deals (a solution is Pareto optimal if neither agent's score can be improved without lowering the other's score). Lower scores indicate inefficient negotiations.

Results are shown in Table \ref{table:endtoend}.
Firstly, we see that the \reinforce and \rollouts models achieve significantly better results when negotiating with the \likelihood model, particularly the \rlrollouts model. 
The percentage of Pareto optimal solutions also increases, showing a better exploration of the solution space.
Compared to human-human negotiations (Table \ref{tab:data}), the best models achieve a higher agreement rate, better scores, and similar Pareto efficiency.
This result confirms that attempting to maximise reward can outperform simply imitating humans.

Similar trends hold in dialogues with humans, with goal-based reasoning outperforming imitation learning. 
The \rollouts model achieves comparable scores to its human partners, and the \rlrollouts model actually achieves higher scores.
However, we also find significantly more cases of the goal-based models failing to agree a deal with humans---largely a consequence of their more aggressive negotiation tactics (see \S\ref{section:analysis}).



\section{Analysis}
\label{section:analysis}
Table \ref{table:endtoend} shows large gains from goal-based methods. In this section, we explore the strengths and weaknesses of our models.
\paragraph{Goal-based models negotiate harder.}
The \rlrollouts model has much longer dialogues with humans than \likelihood (7.2 turns vs. 5.3 on average), indicating that the model is accepting deals less quickly, and negotiating harder. 

A negative consequence of this more aggressive negotiation strategy is that humans were more likely to walk away with no deal, which is reflected in the lower agreement rates. Even though failing to agree was worth 0 points, people often preferred this course over capitulating to an uncompromising opponent---a factor not well captured by the simulated partner in reinforcement learning training or rollouts (as reflected by the larger gains from goal-based models in dialogues with the \likelihood model). 
In particular, the goal-based models are prone to simply rephrasing the same demand each turn, which is a more effective strategy against the \likelihood model than humans.
Future work should address this issue. 

Figure \ref{figure:stubborn} shows an example of our goal-based model stubbornly negotiating until it achieves a good outcome.


\paragraph{Models learn to be deceptive.}
Deception can be an effective negotiation tactic. We found numerous cases of our models initially feigning interest in a valueless item, only to later `compromise' by conceding it. Figure \ref{figure:lying} shows an example.

\newcommand{\itemvalue}[3]{#2x\textbf{#1} \textit{value}=#3}
\newcommand{\deal[3]}


\begin{figure}
\centering
\scalebox{0.70}{%

\begin{tabular}{|llc|}
\hline
\textbf{Input} &&\\
Human & \multicolumn{2}{l|}{\itemvalue{book}{1}{3}  \itemvalue{hat}{1}{1}  \itemvalue{ball}{3}{2}}  \\
\rlrollouts& \multicolumn{2}{l|}{\itemvalue{book}{1}{6}  \itemvalue{hat}{1}{4}  \itemvalue{ball}{3}{0} }\\
\hline
\rlrollouts    &  \multicolumn{2}{l|}{You can have the balls if i can have the hat and book} \\
Human &  \multicolumn{2}{l|}{Nope, give me the books and you can have}\\ 
 &  \multicolumn{2}{l|}{everything else}  \\ 
\rlrollouts    &  \multicolumn{2}{l|}{Ok. I will take the book and one hat} \\
Human & \multicolumn{2}{l|}{ I get all the balls?} \\ 
\rlrollouts    &  \multicolumn{2}{l|}{Yes. I can do that.} \\
Human & \multicolumn{2}{l|}{Ok, deal} \\
\hline
\textbf{Output} & & \textbf{Reward}\\
\rlrollouts & 1x\textbf{book} 1x\textbf{hat} & 10/10\\
Human &  3x\textbf{ball} & 6/10 \\
\hline
\end{tabular}
}

\caption{\label{figure:stubborn}Example of a model stubbornly holding out for maximum points in a dialogue.
}
\end{figure}

\begin{figure}
\centering
\scalebox{0.70}{%

\begin{tabular}{|llc|}
\hline
\textbf{Input} &&\\
Human & \multicolumn{2}{l|}{\itemvalue{book}{3}{2}  \itemvalue{hat}{3}{1}  \itemvalue{ball}{1}{1}}  \\
\rlrollouts& \multicolumn{2}{l|}{\itemvalue{book}{3}{1}  \itemvalue{hat}{3}{2}  \itemvalue{ball}{1}{1} }\\
\hline
Human &  \multicolumn{2}{l|}{I'd like the books and the hats.}  \\ 
\rlrollouts    &  \multicolumn{2}{l|}{I will take the books and hats} \\
Human & \multicolumn{2}{l|}{ I really need the books and at least one other object.} \\ 
\rlrollouts    &  \multicolumn{2}{l|}{Ok, you can have one book and one ball } \\
Human & \multicolumn{2}{l|}{If I can have all the books, I can leave you the rest.} \\
\rlrollouts    &  \multicolumn{2}{l|}{Deal} \\

\hline
\textbf{Output} & & \textbf{Reward}\\
\rlrollouts & 3x\textbf{hat} 1x\textbf{ball} & 7/10\\
Human &  3x\textbf{book} & 6/10 \\
\hline
\end{tabular}
}

\caption{\label{figure:stubborn2}Example of model compromising. 
}

\end{figure}

\paragraph{Models produce meaningful novel sentences.}
One interesting question is whether our models are capable of generating novel sentences in the new circumstances they find themselves in, or if they simply repeat messages from the training data verbatim.
We find that 76\% of messages produced by the \likelihood model in self-play were found in the training data.
We manually examined the novel utterances produced by our model, and found that the overwhelming majority were fluent English sentences in isolation---showing that the model has learnt a good language model for the domain (in addition to results that show it uses language effectively to achieve its goals).
These results suggest that although neural models are prone to the safer option of repeating sentences from training data, they are capable of generalising when necessary. Future work should choose domains that force a higher degree of diversity in utterances.
\begin{figure}[t]
\centering
\scalebox{0.70}{%
\begin{tabular}{|llc|}
\hline
\textbf{Input} &&\\
\rlrollouts & \multicolumn{2}{l|}{\itemvalue{book}{4}{0}  \itemvalue{hat}{2}{5}  \itemvalue{ball}{1}{0}}  \\
Human& \multicolumn{2}{l|}{\itemvalue{book}{4}{1}  \itemvalue{hat}{2}{2}  \itemvalue{ball}{1}{2} }\\
\hline
\rlrollouts    &  \multicolumn{2}{l|}{You can have the ball if I can have everything else.} \\
Human &  \multicolumn{2}{l|}{No can't do that the ball is not going to give }\\ 
 &  \multicolumn{2}{l|}{me anything}  \\ 
\rlrollouts    &  \multicolumn{2}{l|}{I need the hats and you can have the rest} \\
Human & \multicolumn{2}{l|}{ Deal} \\ 
\rlrollouts    &  \multicolumn{2}{l|}{Great! Thank you!} \\
\hline
\textbf{Output} & & \textbf{Reward}\\
\rlrollouts & 2x\textbf{hat} & 10/10\\
Human &  4x\textbf{book} 1x\textbf{ball} & 6/10 \\
\hline
\end{tabular}
}

\caption{\label{figure:lying}Dialogue in which the model's initial interest in the valueless books allows it to compromise while achieving a maximum score.
}

\end{figure}

\paragraph{Maintaining multi-sentence coherence is challenging.} 
One common linguistic error we see \rlrollouts make is to start a message by indicating agreement (e.g. \emph{I agree} or \emph{Deal}), but then going on to propose a counter offer---a behaviour that human partners found frustrating. One explanation is that the model has learnt that in the supervised data, messages beginning with \emph{I agree} are often at the end of the dialogue, and partners rarely reply with further negotiation---so the models using rollouts and reinforcement learning believe this tactic will help their offer to be accepted.



\section{Related Work}

    Most work on goal orientated dialogue systems has assumed that state representations are annotated in the training data \cite{williams:2007,henderson:2014,wen:2016}. 
    The use of state annotations allows a cleaner separation of the reasoning and natural language aspects of dialogues, but our end-to-end approach makes data collection cheaper and allows tasks where it is unclear how to annotate state.
    \newcite{bordes:2016} explore end-to-end goal orientated dialogue with a supervised model---we show improvements over supervised learning with goal-based training and decoding.
    Recently, \newcite{he:2017} use task-specific rules to combine the task input and dialogue history into a more structured state representation than ours.
    
    %

Reinforcement learning (RL) has been applied in many dialogue settings.
RL has been widely used to improve dialogue managers, which manage transitions between dialogue states \cite{singh:2002,pietquin:2011,rieser:2011,gavsic:2013,fatemi:2016}. In contrast, our end-to-end approach has no explicit dialogue manager. 
\newcite{li:2016} improve metrics such as diversity for non-goal-orientated dialogue using RL, which would make an interesting extension to our work.
\newcite{das:2017} use reinforcement learning to improve cooperative bot-bot dialogues.
RL has also been used to allow agents to invent new languages \cite{das:2017,mordatch:2017}.
To our knowledge, our model is the first to use RL to improve the performance of an end-to-end goal orientated dialogue system in dialogues with humans.

    Work on learning end-to-end dialogues has concentrated on `chat' settings, without explicit goals \cite{ritter:2011,vinyals:2015,li:2015}.
    These dialogues contain a much greater diversity of vocabulary than our domain, but do not have the challenging adversarial elements.
    Such models are notoriously hard to evaluate \cite{liu:2016}, because the huge diversity of reasonable responses, whereas our task has a clear objective. Our end-to-end approach would also be much more straightforward to integrate into a general-purpose dialogue agent than one that relied on annotated dialogue states \cite{dodge:2016}.

    There is a substantial literature on multi-agent bargaining in game-theory, e.g. \newcite{nash:1950}. There has also been computational work on modelling negotiations \cite{baarslag:2013}---our work differs in that agents communicate in unrestricted natural language, rather than pre-specified symbolic actions, and our focus on improving performance relative to humans rather than other automated systems. 
    Our task is based on that of \newcite{devault:2015}, who study natural language negotiations for pedagogical purposes---their version includes speech rather than textual dialogue, and embodied agents, which would make interesting extensions to our work.
    The only automated natural language negotiations systems we are aware of have first mapped language to domain-specific logical forms, and then focused on choosing the next dialogue act \cite{rosenfeld:2014,cuayahuitl:2015,keizer:2017}. 
    Our end-to-end approach is the first to to learn comprehension, reasoning and generation skills in a domain-independent data driven way.
    
    Our use of a combination of supervised and reinforcement learning for training, and stochastic rollouts for decoding, builds on strategies used in game playing agents such as AlphaGo \cite{silver:2016}. Our work is a step towards real-world applications for these techniques. Our use of rollouts could be extended by choosing the other agent's responses based on sampling, using Monte Carlo Tree Search (MCTS) \cite{kocsis:2006}. However, our setting has a higher branching factor than in domains where MCTS has been successfully applied, such as Go \cite{silver:2016}---future work should explore scaling tree search to dialogue modelling.

\section{Conclusion}
We have introduced end-to-end learning of natural language negotiations as a task for AI, arguing that it challenges both linguistic and reasoning skills while having robust evaluation metrics.
We gathered a large dataset of human-human negotiations, which contain a variety of interesting tactics.
We have shown that it is possible to train dialogue agents end-to-end, but that their ability can be much improved by training and decoding to maximise their goals, rather than likelihood.
There remains much potential for future work, particularly in exploring other reasoning strategies, and in improving the diversity of utterances without diverging from human language.
We will also explore other negotiation tasks, to investigate whether models can learn to share negotiation strategies across domains.
\section*{Acknowledgments} We would like to thank Luke Zettlemoyer and the anonymous EMNLP reviewers for their insightful comments, and the Mechanical Turk workers who helped us collect data.

\bibliography{emnlp2017}
\bibliographystyle{emnlp_natbib}

\end{document}